\documentclass[conference]{IEEEtran}

\usepackage[noadjust]{cite} 

\usepackage[pdftex]{graphicx}
\DeclareGraphicsExtensions{.pdf,.jpeg,.png}

\usepackage{amsmath}
\usepackage{array}

\usepackage{url}
\usepackage{hyperref} 
\hypersetup{
    colorlinks=true,
    linkcolor=blue,
    filecolor=magenta,
    urlcolor=cyan,
    citecolor=green,
}

\usepackage{fancyhdr}
\usepackage{hyperref} 

\usepackage{booktabs} 
\usepackage{verbatim} 
\usepackage{lipsum} 

\begin{document}
%
\title{BlackBoxToBlueprint: Extracting Interpretable Logic from Legacy Systems using Reinforcement Learning and Counterfactual Analysis}

\author{\IEEEauthorblockN{Vidhi Rathore\IEEEauthorrefmark{1}}
\IEEEauthorblockA{\IEEEauthorrefmark{1}IIIT Hyderabad\\
Hyderabad, India \\ 
Email: vidhi.rathore@research.iiit.ac.in}}


\maketitle
\begin{abstract}
Modernizing legacy software systems is a critical but challenging task, often hampered by a lack of documentation and understanding of the original system's intricate decision logic. Traditional approaches like behavioral cloning merely replicate input-output behavior without capturing the underlying intent. This paper proposes a novel pipeline to automatically extract interpretable decision logic from legacy systems treated as black boxes. The approach uses a Reinforcement Learning (RL) agent to explore the input space and identify critical decision boundaries by rewarding actions that cause meaningful changes in the system's output. These counterfactual state transitions, where the output changes, are collected and clustered using K-Means. Decision trees are then trained on these clusters to extract human-readable rules that approximate the system's decision logic near the identified boundaries. I demonstrated the pipeline's effectiveness on three dummy legacy systems with varying complexity, including threshold-based, combined-conditional, and non-linear range logic. Results show that the RL agent successfully focuses exploration on relevant boundary regions, and the extracted rules accurately reflect the core logic of the underlying dummy systems, providing a promising foundation for generating specifications and test cases during legacy migration.
\end{abstract}


%
\IEEEpeerreviewmaketitle

\section{Introduction}
Legacy software systems, while often critical to business operations, pose significant challenges for maintenance and evolution \cite{Bisbal1999}. Their complex, poorly documented, and potentially outdated codebases make modernization efforts risky and expensive \cite{Sneed2004}. A major challenge is understanding the embedded business logic – the "what" and "why" behind the system's decisions, not just the "how" of its implementation. Simply replicating the observable input-output behavior (behavioral cloning) is insufficient, as it fails to capture the underlying intent and may propagate hidden flaws or obsolete logic into the new system \cite{Wu2019}.

Extracting the core decision logic is crucial for generating accurate specifications for the replacement system, designing effective test cases, and ensuring a successful migration \cite{Canfora2005}. However, manually analyzing large, complex legacy systems is often infeasible. Existing automated program analysis techniques may struggle with missing source code, obsolete languages, or tightly coupled components.

To address this challenge, I propose a pipeline that treats the legacy system as a black box and leverages Reinforcement Learning (RL) to actively discover its decision boundaries. My core idea is to train an RL agent not to mimic the system, but to explore its input space specifically searching for perturbations that cause the output to change significantly. These points represent the system's decision boundaries. I collected these \textbf{counterfactual} transition points, clustered them to identify distinct logical patterns, and then used interpretable models like Decision Trees to translate these patterns into human-readable rules.

This paper makes the following contributions:
\begin{itemize}
    \item A novel pipeline combining RL-based exploration, counterfactual trajectory collection, clustering, and rule extraction for uncovering legacy system logic.
    \item Demonstration of the pipeline on three representative dummy legacy systems with different logical structures.
    \item Analysis of the extracted rules and cluster visualizations, showing their correspondence to the known ground-truth logic of the dummy systems.
    \item Discussion of the potential applications of the extracted logic for legacy system migration tasks like specification generation and testing.
\end{itemize}

The remainder of this paper is structured as follows: Section \ref{sec:methodology} details the proposed pipeline. Section \ref{sec:setup} describes the experimental setup, including the dummy systems and RL parameters. Section \ref{sec:results} presents and discusses the results for each system. Section \ref{sec:conclusion} concludes the paper and outlines future work.

\section{Methodology}
\label{sec:methodology}
My pipeline aims to reverse-engineer the decision logic of a legacy system by observing its input-output behavior under targeted exploration. The process involves the following steps, illustrated conceptually in the sections. 

\fancyhf{} 
\renewcommand{\headrulewidth}{0pt} 
\renewcommand{\footrulewidth}{0pt} 

\thispagestyle{fancy}
\fancyhf{} 
\fancyfoot[L]{\small Code available at: \href{https://github.com/vidhirathore/BlackBoxToBlueprint}{github.com/vidhirathore/BlackBoxToBlueprint}}

\subsection{Legacy System Black-Box Wrapping}
The first step is to treat the legacy system $f$ as a black box. I create a wrapper that allows programmatic interaction, accepting an input vector $x$ and returning the corresponding output $y = f(x)$. This wrapper abstracts away the internal implementation details, which might involve interacting with Command Line Interfaces (CLIs), APIs, or even UI automation tools \cite{Berardi2003}. For this work, I use simple Python functions as stand-ins for wrapped legacy systems.

\subsection{RL Agent for Boundary Exploration}
Instead of passively observing or exhaustively sampling, I employ an RL agent to intelligently explore the input space and identify regions where the system's behavior changes. I model this as an RL problem within a custom environment based on the Gymnasium toolkit \cite{Towers2023}:
\begin{itemize}
    \item \textbf{State ($s$):} The current input vector $x$ being presented to the legacy system. The state space is typically a continuous N-dimensional box defined by the input bounds.
    \item \textbf{Action ($a$):} A perturbation vector $\Delta x$ applied to the current state (input). The action results in a new input $x' = x + \Delta x$, potentially clamped within the input bounds. The action space is also a continuous box, often scaled to encourage small perturbations.
    \item \textbf{Reward ($r$):} The agent is rewarded for discovering decision boundaries. A significant positive reward (e.g., +1) is given if the legacy system's output for the next state $x'$ differs from the output for the current state $x$ ($f(x') \neq f(x)$). Otherwise, the reward is zero or a small negative value (to encourage efficiency, though zero was used here).
    \item \textbf{Episode Termination:} An episode ends after a fixed number of steps or potentially if the agent reaches a specific (though often undefined) goal state. Here, I use a fixed step limit.
\end{itemize}
I utilize the Proximal Policy Optimization (PPO) algorithm \cite{Schulman2017}, a state-of-the-art on-policy RL algorithm implemented in Stable-Baselines3 \cite{stable-baselines3}, known for its stability and performance across various environments. The agent learns a policy $\pi(a|s)$ that maximizes the expected cumulative reward, effectively learning to perturb the input state $x$ in ways that are likely to trigger a change in the output $y$.

\subsection{Counterfactual Trajectory Generation}
Once the RL agent is trained, I deploy it in the environment to generate trajectories. I are specifically interested in the transitions where the reward was positive, indicating a change in the legacy system's output. I log these counterfactual transitions as tuples: $[s, a, s', y_{prev}, y_{curr}, r]$, where $s=x$, $s'=x'$, $y_{prev}=f(x)$, $y_{curr}=f(x')$, and $r>0$. These logged transitions represent points in the input space situated just before a decision boundary was crossed.

\subsection{Clustering of Decision Patterns}
The collected counterfactual states ($s$ from the logged transitions) likely form groups corresponding to different decision boundaries or regions of the legacy logic. To identify these patterns, I apply a clustering algorithm to the set of collected states $\{s | r>0\}$. I use K-Means \cite{MacQueen1967}, a simple and widely used partitioning algorithm, implemented in Scikit-learn \cite{scikit-learn}. K-Means aims to partition the $N$ states into $k$ clusters $C = \{C_1, C_2, ..., C_k\}$ so as to minimize the within-cluster sum of squares. Each cluster $C_i$ represents a group of input states that were similarly situated just before triggering an output change. For systems with 2D input, I visualize these clusters and their centroids.

\subsection{Rule Extraction from Clusters}
The final step is to translate the identified clusters into interpretable rules that approximate the legacy system's decision logic near the boundaries. I employ Decision Trees \cite{Quinlan1986} for this purpose, again using the Scikit-learn implementation. I train a Decision Tree classifier where the input features are the counterfactual states $s$, and the target variable is the cluster label assigned by K-Means. The resulting tree partitions the input space based on feature thresholds. By limiting the tree's depth for interpretability, the paths from the root to the leaves can be converted into simple IF-THEN rules (e.g., using \texttt{export\_text}) that describe the input conditions leading to different clusters of boundary-crossing behavior. These rules serve as approximations of the legacy system's underlying logic in those critical regions.

\section{Experimental Setup}
\label{sec:setup}

\subsection{Dummy Legacy Systems}
To evaluate my pipeline, I implemented three dummy legacy systems in Python, representing different types of decision logic:

\begin{itemize}
    \item \textbf{System 1 (Threshold):} Takes a 1D input $x_0$. Outputs 'Category A' if $x_0 \le 5.0$, 'Category B' otherwise. Input bounds: [-10.0, 10.0].
    \item \textbf{System 2 (Combined):} Takes a 2D input $(x_0, x_1)$. Outputs 'High' if $x_0 > 0$ and $x_1 > 0$, 'Low' if $x_0 < 0$ and $x_1 < 0$, and 'Medium' otherwise. Input bounds: [-5.0, 5.0] for both dimensions.
    \item \textbf{System 3 (Nonlinear):} Takes a 1D input $x_0$. Outputs integer score 10 if $-2 < x_0 < 2$, score 20 if $x_0 \ge 2$ or $x_0 \le -2$, and 0 otherwise (though 0 is unreachable with these bounds). Input bounds: [-5.0, 5.0].
\end{itemize}
These systems provide known ground truth against which I can validate the extracted logic.

\subsection{RL Agent Configuration}
For all experiments, I used the PPO algorithm from Stable-Baselines3 with the default \texttt{MlpPolicy} (a multi-layer perceptron actor-critic network). Key parameters were:
\begin{itemize}
    \item \textbf{Total Training Timesteps:} 20,000
    \item \textbf{Environment Wrapper:} \texttt{make\_vec\_env} with \texttt{n\_envs=4} (4 parallel environments for faster training).
    \item \textbf{Learning Rate:} Default (0.0003).
    \item \textbf{Gym Environment:} \texttt{LegacyExplorerEnv} with \texttt{max\_steps=100} per episode during training and \texttt{action\_scale=1.0}. TensorBoard logging was enabled.
\end{itemize}
Models were trained on a machine with an NVIDIA GeForce RTX 3050 (4GB VRAM) GPU using CUDA 12.4, although PPO primarily utilized CPU computation for MLP policies. Training was performed once per system, and the trained agent was saved for subsequent counterfactual generation.

\subsection{Analysis Configuration}
After training, the agent was loaded and run for analysis:
\begin{itemize}
    \item \textbf{Trajectory Collection:} 100 episodes were run using the deterministic policy (\texttt{agent.predict(obs, deterministic=True)}) with \texttt{max\_steps=200} per episode. Counterfactual transitions were saved to CSV files.
    \item \textbf{Clustering:} K-Means was applied to the collected counterfactual 'state' vectors with \texttt{n\_clusters=4} and \texttt{n\_init=10}.
    \item \textbf{Rule Extraction:} A Decision Tree Classifier was trained on the states with cluster labels as targets. \texttt{max\_depth} was set to $\max(3, \mathrm{input\_dim} + 1)$. Rules were extracted using \texttt{export\_text}.
\end{itemize}

\section{Results and Discussion}
\label{sec:results}
I now present the results of applying the pipeline to each dummy legacy system.

\subsection{System 1: Threshold Logic}
\begin{itemize}
    \item \textbf{Counterfactuals Found:} 75 transitions where the output changed from 'Category A' to 'Category B' or vice-versa were collected. Examination of the \texttt{system\_1\_threshold\_trajectories.csv} file confirms these transitions occur when the \texttt{next\_state} crosses the 5.0 threshold (e.g., state 4.93 -> \texttt{next\_state} 5.04).
    \item \textbf{Clustering Visualization:} Skipped, as expected for 1D input data.
    \item \textbf{Extracted Rules:} The decision tree rules attempt to separate the 4 clusters based on \texttt{input\_0}:
    \begin{verbatim}
|--- input_0 <= 4.96
|   |--- input_0 <= 4.92
|   |   |--- input_0 <= 4.89
|   |   |   |--- class: Cluster_3
|   |   |--- input_0 >  4.89
|   |   |   |--- class: Cluster_1
|   |--- input_0 >  4.92
|   |   |--- class: Cluster_2
|--- input_0 >  4.96
|   |--- class: Cluster_0
    \end{verbatim}
    \item \textbf{Discussion:} The RL agent successfully identified the region around $x_0 = 5.0$ as the critical decision boundary. All 75 counterfactual states are necessarily clustered very close to this value. The decision tree's split points (4.96, 4.92, 4.89) clearly reflect this proximity to the true threshold of 5.0. While the specific cluster assignments might seem arbitrary given the single boundary, the *location* of the splits extracted by the tree accurately pinpoints the legacy system's core logic: a threshold exists near $x_0 = 5.0$. This rule is directly usable for specifying the new system's behavior.
\end{itemize}

\subsection{System 2: Combined Conditions Logic}
\begin{itemize}
    \item \textbf{Counterfactuals Found:} 200 transitions were collected (based on examining \texttt{system\_2\_combined\_trajectories.csv}, actual count may vary slightly per run). These transitions correspond to crossing the $x_0=0$ axis, the $x_1=0$ axis, or both near the origin, leading to changes between 'High', 'Medium', and 'Low' outputs.
    \item \textbf{Clustering Visualization:} The 2D cluster plot was generated (Fig. \ref{fig:sys2_clusters}).
        \begin{figure}[!ht]
        \centering
        \includegraphics[width=0.9\columnwidth]{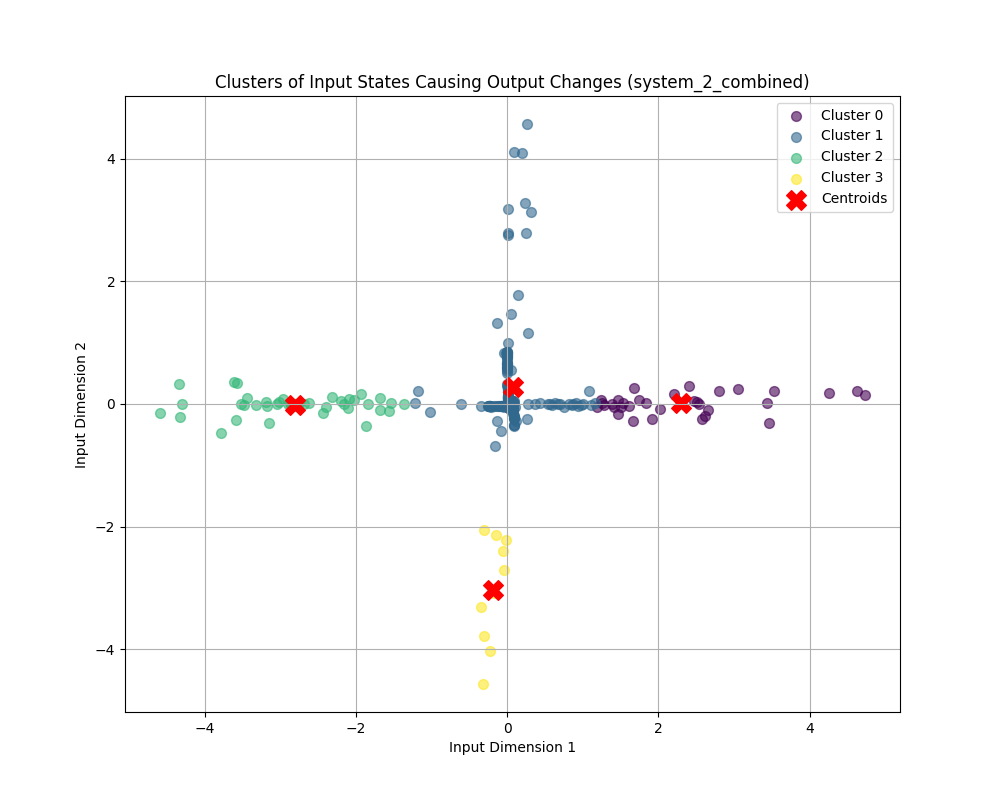} 
        \caption{Clusters of Input States Causing Output Changes for System 2 (Combined Conditions). Points represent states just before an output change. Red 'X' markers indicate cluster centroids.}
        \label{fig:sys2_clusters}
        \end{figure}
    \item \textbf{Extracted Rules:} The decision tree rules use both \texttt{input\_0} and \texttt{input\_1}:
    \begin{verbatim}
|--- input_0 <= -1.29
|   |--- class: Cluster_2
|--- input_0 >  -1.29
|   |--- input_0 <= 1.18
|   |   |--- input_1 <= -1.37
|   |   |   |--- class: Cluster_3
|   |   |--- input_1 >  -1.37
|   |   |   |--- class: Cluster_1
|   |--- input_0 >  1.18
|   |   |--- class: Cluster_0
    \end{verbatim}
    \item \textbf{Discussion:} The cluster plot (Fig. \ref{fig:sys2_clusters}) visually confirms that the RL agent focused its exploration near the axes ($x_0=0$ and $x_1=0$). The four clusters roughly correspond to regions near the positive x-axis (Cluster 0, purple), negative x-axis (Cluster 2, green), positive y-axis (Cluster 1, blue), and negative y-axis (Cluster 3, yellow), specifically where crossing an axis changes the output category. The centroids (Red 'X') lie near these boundary regions. The decision tree rules capture this logic, although the thresholds (-1.29, 1.18 for \texttt{input\_0}; -1.37 for \texttt{input\_1}) are not exactly zero. This is likely due to the agent receiving rewards for any change, the specific distribution of collected points, and the nature of K-Means/Decision Tree approximations. However, the rules clearly identify that conditions on both \texttt{input\_0} and \texttt{input\_1} around zero determine the outcome changes. Cluster 1 (blue, large concentration near origin) likely captures many transitions crossing either axis close to (0,0). The rules suggest: crossing the negative x-axis (Cluster 2), crossing the negative y-axis (Cluster 3), crossing the positive x-axis (Cluster 0), or being near the origin but not satisfying the other conditions (Cluster 1). This provides valuable insight into the structure (\texttt{x} vs \texttt{y} boundaries) of the legacy logic.
\end{itemize}

\subsection{System 3: Nonlinear Ranges Logic}
\begin{itemize}
    \item \textbf{Counterfactuals Found:} 65 transitions were collected where the output changed between 10 and 20. These correspond to crossing the boundaries at $x_0 = -2.0$ and $x_0 = 2.0$.
    \item \textbf{Clustering Visualization:} Skipped (1D input). 
    \item \textbf{Extracted Rules:} The decision tree rules identify two main splitting regions:
    \begin{verbatim}
|--- input_0 <= 0.00
|   |--- input_0 <= -2.08
|   |   |--- class: Cluster_3
|   |--- input_0 >  -2.08
|   |   |--- class: Cluster_1
|--- input_0 >  0.00
|   |--- input_0 <= 2.06
|   |   |--- class: Cluster_0
|   |--- input_0 >  2.06
|   |   |--- class: Cluster_2
    \end{verbatim}
    \item \textbf{Discussion:} The pipeline successfully identified the two critical boundaries of the legacy system. The decision tree splits occur at \texttt{-2.08} and \texttt{2.06}, which are very close approximations of the true boundaries at -2.0 and 2.0. The intermediate split at \texttt{0.00} simply divides the space between the two main boundaries. The rules clearly map to the underlying logic: states below -2.08 (Cluster 3) and above 2.06 (Cluster 2) represent inputs just outside the central range, while states between -2.08 and 2.06 (Clusters 1 and 0) represent inputs just inside or crossing into the central range. This accurately reflects the "if $-2 < x_0 < 2$ then 10, else 20" logic near the transition points.
\end{itemize}

\subsection{Implications for Legacy Migration}
The results across all three systems demonstrate the pipeline's potential. By focusing on counterfactual transitions, the RL agent efficiently locates decision boundaries. Clustering groups similar transition types, and the decision tree provides interpretable rules summarizing the logic at these boundaries. This extracted logic can directly inform migration efforts:
\begin{itemize}
    \item \textbf{Specification:} Rules like "\texttt{input\_0 <= ~5.0 -> Cat A}" or "\texttt{input\_0 > ~2.0 or input\_0 < ~-2.0 -> Score 20}" are far clearer specifications than pointing to legacy code.
    \item \textbf{Testing:} The identified boundary values (-2, 2, 5 for System 1/3; axes near 0 for System 2) are critical values for targeted testing of the new system. Test cases should explicitly cover inputs just below, at, and just above these discovered thresholds.
    \item \textbf{Understanding:} The process provides insight into *what* conditions trigger behavior changes, aiding overall comprehension of the legacy system's function.
\end{itemize}
While demonstrated on simple systems, this approach offers a scalable way to tackle more complex black boxes where manual analysis is impractical. The main limitations currently are the reliance on dummy systems and the potential complexity of rules extracted from systems with very high-dimensional inputs or intricate boundaries.

\section{Conclusion and Future Work}
\label{sec:conclusion}
This paper presented a pipeline utilizing Reinforcement Learning, counterfactual analysis, clustering, and decision tree induction to extract interpretable decision logic from black-box legacy systems. By training an RL agent to seek out input perturbations that cause output changes, I efficiently identified critical decision boundary regions. Clustering these boundary-adjacent states and subsequently applying decision tree analysis yielded rules that accurately reflected the underlying logic of three distinct dummy legacy systems.

The results suggest this approach is a promising technique for aiding legacy system modernization. The extracted rules can serve as valuable artifacts for generating specifications, guiding development, creating targeted test plans, and improving overall understanding of the legacy system's essential behavior, thereby reducing migration risks.

Future work will focus on several directions:
\begin{itemize}
    \item Applying the pipeline to real-world legacy systems using appropriate wrappers (e.g., API calls, CLI interactions).
    \item Exploring alternative RL algorithms (e.g., SAC for potentially more sample efficiency) and reward shaping strategies.
    \item Investigating more sophisticated clustering (e.g., DBSCAN, HDBSCAN) and rule extraction techniques (e.g., symbolic regression, rule induction algorithms like RuleFit) for potentially more accurate or concise rules, especially for complex boundaries.
    \item Developing methods to handle high-dimensional input spaces, possibly involving dimensionality reduction techniques (e.g., UMAP, PCA) prior to clustering or analysis.
    \item Refining the process for legacy systems with numerical outputs, including defining "meaningful change" thresholds for the reward function.
    \item Evaluating the scalability of the approach on larger, more complex systems.
\end{itemize}
By addressing these areas, I aim to develop a robust and practical toolset for reverse-engineering and understanding the core logic embedded within opaque legacy software.

\bibliographystyle{IEEEtran}

\end{document}